\documentclass{article}

\usepackage{arxiv}

\usepackage[utf8]{inputenc} % allow utf-8 input
\usepackage[T1]{fontenc}    % use 8-bit T1 fonts
\usepackage{hyperref}       % hyperlinks
\usepackage{url}            % simple URL typesetting
\usepackage{booktabs}       % professional-quality tables
\usepackage{amsfonts}       % blackboard math symbols
\usepackage{nicefrac}       % compact symbols for 1/2, etc.
\usepackage{microtype}      % microtypography
\usepackage{lipsum}

\usepackage{graphicx}
\graphicspath{{figures/}}  % Location of 
\usepackage[table,xcdraw]{xcolor}
\usepackage{longtable}
\usepackage[normalem]{ulem}
\useunder{\uline}{\ul}{}
\usepackage{lscape}
\usepackage{longtable}
\usepackage{placeins}
\usepackage{amsmath}
\usepackage{subcaption}
\usepackage{amssymb}
\usepackage{float}
\title{Conditional GAN for timeseries generation}

\author{
  Kaleb ~Smith\\
  Department of Computer Engineering and Sciences\\
  Florida Institute of Technology\\
  Melbourne, FL 32901 \\
  \texttt{ksmith012007@my.fit.edu} \\
   \And
  Anthony O.~Smith \\
  Department of Computer Engineering and Sciences\\
  Florida Institute of Technology\\
  Melbourne, FL 32901 \\
  \texttt{anthonysmith@fit.edu} \\
}

\begin{document}
\maketitle

\begin{abstract}
It is abundantly clear that time dependent data is a vital source of information in the world. The challenge has been for applications in machine learning to gain access to a considerable amount of quality data needed for algorithm development and analysis. Modeling synthetic data using a Generative Adversarial Network (GAN) has been at the heart of providing a viable solution. Our work focuses on one dimensional times series and explores the “few shot” approach, which is the ability of an algorithm to perform well with limited data. This work attempts to ease the frustration by proposing a new architecture, Time Series GAN (TSGAN), to model realistic time series data. We evaluate TSGAN on 70 data sets from a benchmark time series database. Our results demonstrate that TSGAN performs better than the competition both quantitatively using the Fréchet Inception Score (FID) metric, and qualitatively when classification is used as the evaluation criteria.
\end{abstract}

% keywords can be removed
\keywords{Generative Adversarial Networks (GANs) \and Time Series Generation \and Few Shot \and Conditional GANs \and Wasserstein GANs \and Unsupervised Learning }

\section{Introduction}
\label{S:1}
Many prominent domains such as financial, medical, weather, and geo-physics are some of the most relevant sources of time dependent data. With this increase in information, comes the increase in demand to explore machine learning tasks (classification, prediction, detection, etc) to address challenges. Factors such as high collection cost or data quality make the acquisition of the data difficult, to acquire or impossible to use. This limited data drastically impacts the performance of machine learning and impedes the models ability to perform.

To remedy the challenges, we propose a novel architecture called Time Series GAN (TSGAN). TSGAN uses two GANs in unison to model fake time series examples. The popular generative model GAN \cite{Goodfellow2014GenerativeNets}, is an unsupervised deep learning method in which two deep networks are pitted against each other to generate synthetic data. The original GAN was extended to show that minimizing the Wasserstein distance leads to more stable training. The first GAN is responsible for transforming random latent vectors to valid spectrogram images from a distribution. The added second GAN is seeded with the conditional spectrogram images to then generate time series. On 70 widely used data sets, we show that our TSGAN architecture models a variety of data having low to limited training data, yet learns better generators in terms of time series to produce high fidelity and diverse synthetic time series. This is measured numerically by the 1-nearest neighbor test, Fréchet inception distance, the reconstruction error, and qualitatively by visually analyzing the generated samples.  The paper is organized as follows: first, we discuss the related work and show that there is not a copious amount of work in 1D generative models. Second, we describe our TSGAN architecture and explain the training methodology. Third, we expound the experimental setup and both qualitative and quantitative metrics used for evaluation. Finally, we conclude our work and discuss a future path of research interest.

\section{Related Works}
\label{relatedWorks}
Methods used and published in literature have all been either supervised autoregressive models (predicting) or unsupervised GANs (generating).  Though there are other methods out there to generate data through different means, these two methods make up the majority of the literature.  Since our method is modeled after GANs, we will focus on that literature review more in depth than the autoregressive models. 

Autoregressive models are nice for generation networks because of the stability and ability train in parallel, fast inference to the new predictions, and no truncated backpropagation through time.  It is also similar to the recurrent models which have the ability to generate data by taking your input data (or some spread over the input data) and use that to predict the next time step of the data \cite{Bengio2015ScheduledNetworks}. Autoregressive models make predictions when a value from a time series is regressed on previous values from that same time series. The deep autoregressive models use neural networks to learn the function to do the prediction of the future time series data. These predictions can be seen as generating the distribution of the time series based on previous time steps.  

The most well-known deep autoregressive model in literature would be WaveNet \cite{Oord2016PixelNetworks}, which was inspired by PixelCNN \cite{Salimans2017Pixelcnn++:Modifications}.  Both used deep autoregressive networks to produce pixels sequentially for image generation with both an autoregressive recurrent neural network (RNN) and convolutional neural network (CNN).  Where autoregressive networks do well is in the ability to generate both continuous and discrete data, which can not be done with other techniques.  It also allows for training and testing splits in training and validating your networks, which cannot be said for some unsupervised training techniques.  Where autoregressive models fail is their large computational complexity for generating one sample.  For example, WaveNet has $2^{16}$ softmax calculations per time step, which complicates the model to generate a one second audio clips in a day’s training time.  Also, autoregressive models are not truly generative in the sense that they must be externally conditioned \cite{Williams1989ANetworks}.

Compared to the image counterpart \cite{Zhu2017UnpairedNetworks, Wu2018LearningReconstruction, Qiao2019MirrorGAN:Redescription, Brock2018LargeSynthesis, Reed2016GenerativeSynthesis}, the available literature that describes work with time series GANs is meager.  GANs have been an afterthought for time series generation, which shows promise for newer techniques that do not conform to the image GAN architectures. One of the first approaches on the use of GANs for time series generation was done by Morgren \cite{Mogren2016C-RNN-GAN:Training}, utilizing both recurrent neural networks (RNN) and GANs to synthesize music data.  The author shows how a network and adversarial training can be used to train and remain highly flexible and expressive with continuous sequence data for tone length, frequencies, intensities, and timing.  This varies from the original approach of using RNNs through symbolic representations only.  This work failed to span outside of music though, outlaid the groundwork going forward.  Esteban et al. \cite{Esteban2017Real-valuedGans} utilized the work done with RNNs and GANs by using a recurrent conditional generative adversarial network (RCGAN) to generate time series data.  Their paper demonstrated that time series generation was possible with simple sinusoidal data and real world medical data such as oxygen saturation, heart rate monitoring, etc., but failed to give any real plots of the generated data to compare.  Harada et al. \cite{Haradal2018BiosignalNetworks} also looked to generate biosignal data like electroencephalographic(EEG) and electrocardiography(ECG) data using a similar structured RNN GAN.

To build off the work for synthesizing medical data, Hartmann et al. \cite{Hartmann2018EEG-GAN:Signals} chose to focus EEG data utilizing an improved Wasserstien GAN (WGAN) and progressively growing the GAN to generate medical data and not RNN.  This method showed to perform better than a vanilla GAN, however added to the complexity of the training.  Similarly, Smith et al. \cite{Smith2019TimeGAN} used a generic WGAN to generate synthetic time series data by training on the raw wave forms of the data.  Though this method was simple in the approach, it was one of the only papers showing that a simple image approach could be scaled to 1D. Simonetto \cite{Simonetto2018GeneratingTransactions} proposed using a mixture of variational autoencoders (VAE) and WGANs to generate spiking time series data simulating many banking transactions.  When  compared to handcrafted features, a VAE, and a WGAN their results shows little to no advantage.  Alzantot et al. looked to stack multiple methods as well to create their SenseGAN which produces sensor data.  They chose to stack multiple long-short-term-memory (LSTM) networks with Mixture Density Networks (MDN) for their generator model and LSTMs for their discriminator due to the their ability and reputation to handle sequential data.  Their work showed they could fool a discriminator model 50\% of the time but with no visual results or extended work in experiments.

Zhang et al. \cite{Zhang2018GenerativeGrids} proposed using a GAN to learn time series associated with smart grids.  They used GANs to generate the probabilistic features inherent to smart grid time series data.  They then use this data and post process it to create their synthetic time series; so their method does not use GAN to generate the actual time series, but rather the characteristics that make it up.

Ramponi et al. \cite{Ramponi2018T-cgan:Sampling} introduced a method for time series generation using the time steps as conditions on the generated, called time-conditioned GAN (T-CGAN).  This method was shown to increase classification accuracy when the generated time series was used to augment the training data of the original training set.  TSGAN correlates well with this method since we see the benefit of conditioning the generator with better information than just random noise space input.  An interesting network for time series generation  comes from a more recent paper by Brophy et al \cite{Brophy2019QuickGANs}.  When they mapped their time series data to images, applied a GAN to generate more images, and then mapped those synthetic images to create 1D synthesized data.  This proved to be quick and provide comparable results in the time series domains applied that leveraged this approach.

Qualitatively, this paper shows compelling results against other generative models.  However, not one synthetic signal is shown to validate, and the training times proved to be longer than a GAN only approach.  Our method looks to build off of Smith et al. \cite{Smith2019TimeGAN} in their WGAN approach for time series generation while using the concepts from T-CGAN for conditioning the input to the generator.  With their results, we see promising ability to utilize multiple WGANs inside of TSGAN in order to generate realistic synthetic data.

\section{TSGAN}
\label{S:2}

\begin{figure}[t!]
    \centering
    \includegraphics[scale=.5]{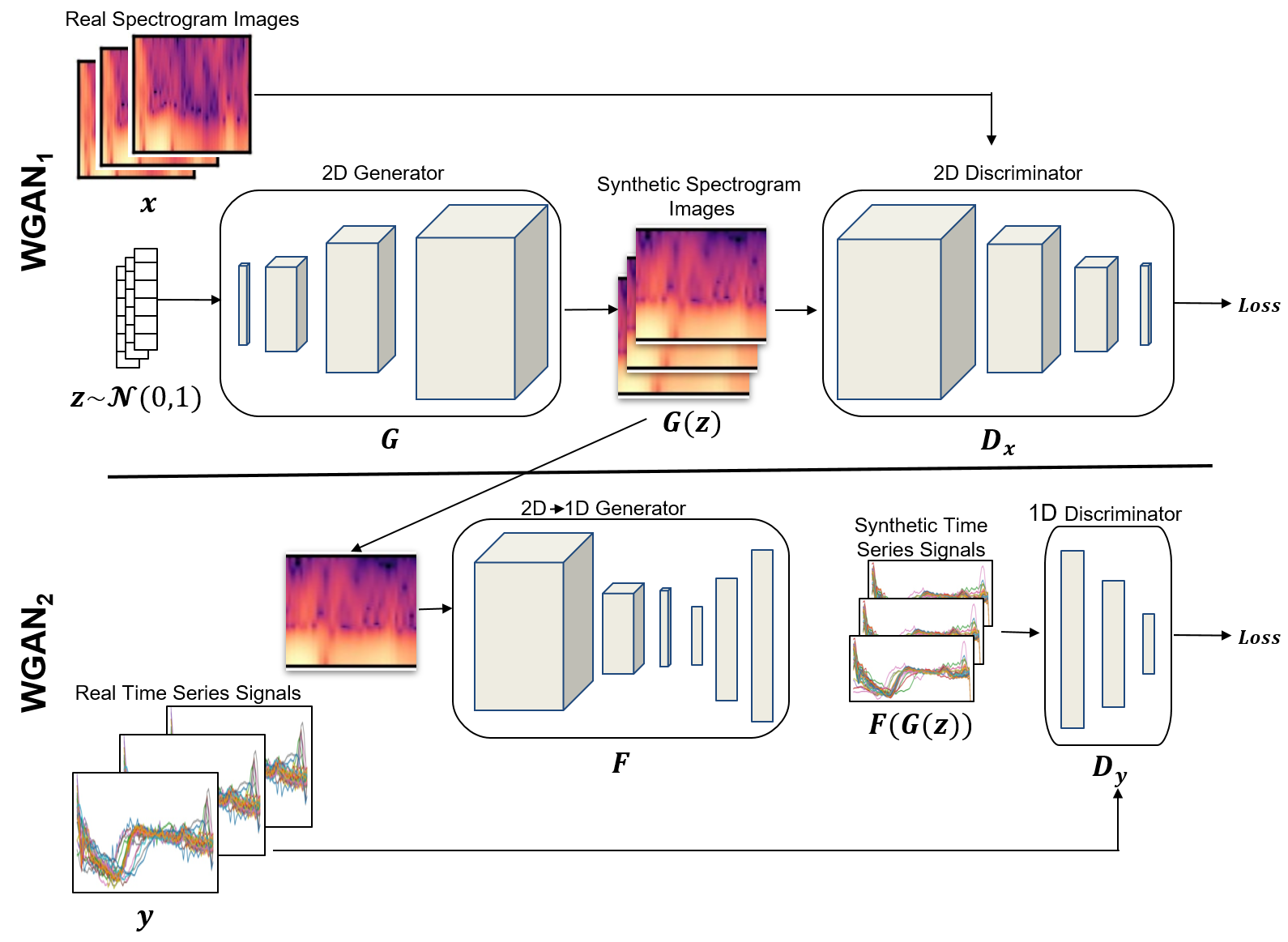}
    \caption{The 2-layer TSGAN architecture for synthetic time series generation. WGAN${_1}$ in layer 1 uses a random vector to produce a  synthetic spectrogram. The layer 2 conditional WGAN${_2}$ converts the synthetic image from layer 1 to generate a synthetic time series.} 
    \label{TSGAN_Arch}
\end{figure}

Our TSGAN architecture can been seen in Figure \ref{TSGAN_Arch} and is built from two WGANs (one regular and one conditional). Since WGANs are used so intensively in our approach, the following will touch on their mathematical principals as well as lay out our cost function for optimization. Then the WGANs in TSGAN will be discussed respectively as well as intuition that follows the methodology. 

WGANs use the Wasserstein distance (or earth mover's distance) which measures the similarities between two probability distributions, which in our case the true distribution and generated distribution of the synthetic data.  Intuitively, the the equation defines the cost needed to convert one data distribution to the other, comparing the similarity between input sample ${x}$ and ${y}$. As shown in 

\begin{equation} \label{EMD}
W(\mathbb{P}_{r}, \mathbb{P}_{\theta}) = \inf_{\lambda} \mathbb{E}_{(x,y)} {||x-y||}, 
\end{equation} 

the distance is defined between a sample ${x}$ from $\mathbb{P}_{r}$ and a sample ${y}$ from $\mathbb{P}_{\theta}$. The measure in Equation \ref{EMD} is the greatest lower bound (${inf}$) of energy required to transport (${\lambda}$) the distribution of the generated data $\mathbb{P}_{\theta}$ to the distribution of the true data $\mathbb{P}_{r}$.  This differs from the original GAN approach of minimizing the Kullback-Leibler (KL) Divergence, which is a measure of how two distributions differ from one another and not a true distance metric \cite{Goodfellow2014GenerativeNets}. The KL-Divergence causes a common problem of vanishing gradients; a training issue when the model distribution
increases in dissimilarity to the true distribution and the generator stops learning i.e the KL-Divergence grows large \cite{GANarjovsky2017wasserstein}. Also of concern with the KL-Divergence is the large variance in the gradients causing training to be unstable \cite{Goodfellow2014GenerativeNets,GANarjovsky2017wasserstein}.

The Wasserstein distance though has smoother gradients everywhere which allows the generator to learn regardless of the outcome of the discriminator. This is to help with vanishing gradients, where the discriminator learns too quickly and in response is not able to train the generator due to the gradients going to zero. WGANs also help with the issue of mode collapse, which is when the variation in generated data is locked to one mode and doesn't learn outside of it. Mode collapse is a challenging issue especially when little data is present for training due to the lack of the variation in the data. It is difficult however to achieve this minimization of the Wasserstein distance unless looked at in its dual form, found by utilizing the Kantorovich-Rubinstein
duality \cite{Gulrajani2017ImprovedGANs}.

\begin{equation} \label{KRD}
\begin{split}
W(\mathbb{P}_{r},\mathbb{P}_{\theta})=\sup_{||f||_{L}\leq K}\mathbb{E}_{x\sim\mathbb{P}_{r}}[f(x)]-\mathbb{E}_{x\sim\mathbb{P}_{\theta}}[f(x)] \\
s.t.\qquad|f(x_{1})-f(x_{2})|\leq|x_{1}-x_{2}|,
\end{split}
\end{equation} 

where $\mathbb{P}_{r}$ is the real distribution, $\mathbb{P}_{\theta}$ is the generated data distribution, and $f$ is a $K$-Lipschitz function following the constraint in Equation \ref{KRD}.  It should be noted that for our work, we let $K=1$ which simplifies the notation of any $K$ value. The minimization problem of the original Wasserstein distance now becomes a maximization problem in this dual form with restrictions that the function ${f}$ remains 1-Lipschitz to ensure the gradients remain smooth, which we will touch on more later in this section. We allow our network to learn this 1-Lipschitz function, ${f}$, by building our network similar to a standard discriminator, however it behaves like a critic. This is because instead of a probability outcome of real versus synthetic, it will give a scalar score rating how well our generator is doing. This difference in output now makes our discriminator a critic in the sense that it will grade the generator on its performance instead of just saying real or not real. Note for simplicity we will continue to call our critic a discriminator. The loss function for WGAN is
\begin{equation}
\min_{G}\max_{D}\mathbb{E}_{x\sim \mathbb{P}_r}[D(x)]-\mathbb{E}_{z\sim \mathcal{N}(z)}[D(G(z))],
\end{equation} where $\mathbb{P}_{r}$ is the real distribution, $\mathbb{P}_{\theta}$ is the generated data distribution, and ${z}$ is a random vector from a Gaussian distribution $\mathcal{N}$. This loss function looks to compare the expectation of the discriminator's score of the original data and the discriminator's score of the generated data to give this generation quality score. By minimizing the output of the generator, ${G}$, the discriminator, ${D}$ is maximized causing an optimal output of realistic synthetic data.   

The gradients for the discriminator and generator are respectively defined as
\begin{equation}
\nabla_{\theta_D}\frac{1}{m}\sum_{i=1}^{m}[f(x^{(i)})-f(G(z^{(i)}))]
\end{equation}

\begin{equation}
\nabla_{\theta_G}\frac{1}{m}\sum_{i=1}^{m}f(G(z^{(i)})),
\end{equation} where ${\nabla_{\theta_D}}$ is the gradient of the discriminator network's parameters and ${\nabla_{\theta_G}}$ is the gradient of the generator network's parameters. The issue with these gradients though is the constraint proposed on the 1-Lipschitz function, $f$, is not enforced just through the gradients, so the original WGAN paper by \cite{GANarjovsky2017wasserstein} proposed gradient clipping, where the weights, $w$, are updated via some optimizer and then clipped to be between two hyperparameter constants (typically -0.01 ans 0.01) .  Though the clipping ensures the transport function $f$ remains 1-Lipschitz, it actually causes more instability issues to the training and does not solve the mode collapse problem. The main issue being that clipping is highly sensitive to the values in which is chosen to clip between and has the likelihood of not allowing convergence or good results. This issue was addressed in \cite{GANgulrajani2017improvedWGAN} where the solution to this clipping was to penalize the gradients to impose the 1-Lipschitz constraint on $f$. This requires us to redefine our loss function to one with a regularization term.
\begin{equation} \label{lossFunctionWGAN}
L=\mathbb{E}_{\boldsymbol{x}\sim\mathbb{P}_{r}}[D(\boldsymbol{x})]-\mathbb{E}_{\boldsymbol{\tilde{x}}\sim\mathbb{P}_{\theta}}[D(\boldsymbol{\tilde{x}})] \\    + \lambda\mathbb{E}_{\boldsymbol{\hat{x}}\sim\mathbb{P}_{\hat{x}}}[(||\nabla_{\boldsymbol{\hat{x}}}D(\boldsymbol{\hat{x}})||_{2}-1)^{2}]
\end{equation} Equation \ref{lossFunctionWGAN} where $\mathbb{P}_{r}$ is the real distribution, $\mathbb{P}_{\theta}$ is the generated data distribution, ${x\tilde{}}$ is a sample from the generated distribution, $\mathbb{P}_{x\hat{}}$ is the sampling distribution uniformly sampled on a straight line between $\mathbb{P}_{r}$ and  $\mathbb{P}_{\theta}$, ${x\hat{}}$ is one of those samples from the sampling distribution, $\lambda$ is a strength term on the gradient penalty, and $\nabla_{\boldsymbol{\hat{x}}}D(\boldsymbol{\hat{x}})$ is the gradient of the discriminator on the sample  ${x\hat{}}$. This WGAN-GP is used on both TSGAN's networks and generates robust results without mode collapse or vanishing gradients. For simplicity, we will continue to call our WGAN-GP portions of our model WGAN. This loss function ensures the generator will learn even when the discriminator is performing its job well, and shows to combat mode collapse on experimental results. The loss WGAN loss function also allows for the optimization to be over image quality instead of just real or synthetic, allowing for better learning and data generation. Again, TSGAN is made up of two WGANs, the first is a regular WGAN and the second being a conditional WGAN. 

\subsection{TSGAN System Architecture}
We refer back to the TSGAN architecture shown in Figure \ref{TSGAN_Arch}.  Here, WGAN${_1}$ with generator, $G$, is a convolutional based generator that transforms a single random vector, $z$ to a RGB spectrogram image, ${G(z)}$. This $G$ contains convolutional layers and Leaky ReLU activation functions which is based on the work by Gulrajani's \cite{Gulrajani2017ImprovedGANs} that improved the WGAN. These synthetic spectrograms generated by TSGAN's first WGAN are essentially synthetic power spectral densities of time series data not yet synthesized. It is this spectrogram data that seeds the the conditional WGAN generator and attempts to find through conditional learning the ideal timeseries.

WGAN${_1}$ is optimized using Adam and is trained independently, optimizing 

\begin{equation}
\label{lossFunctionWGAN_1}
L=\mathbb{E}_{\boldsymbol{x}\sim\mathbb{P}_{r}}[D_x(\boldsymbol{x})]-\mathbb{E}_{\boldsymbol{z\sim\mathcal{N}(0,1)}}[D(G(z))] +\lambda\mathbb{E}_{\boldsymbol{\hat{x}}\sim\mathbb{P}_{\hat{x}}}[(||\nabla_{\boldsymbol{\hat{x}}}D(\boldsymbol{\hat{x}})||_{2}-1)^{2}].
\end{equation}

Our second WGAN${_2}$ with generator $F$, receives the 2D spectrogram images, $G(z)$ generated by $G$ and outputs a timeseries, $F(G(z))$. We believe this output from $F$ is leveraging the frequency heat map transition to gather more in depth conditioned information to generate better time series data. Our conditional generator, $F$, is structured similar to a 2D encoder network that then compresses the information of the spectrogram image to a single layer in which it then decode into 1D signals. The encoder is made of convolutional layers with strided convolutions which are used so the network can learn its own spatial downsampling \cite{Radford2015UnsupervisedNetworks}. It's a known fact that convolutional layers in a deep network becomes bias to texture based features \cite{Ford2019AdversarialNoise, Geirhos2018ImageNet-trainedRobustness, Hendrycks2019BenchmarkingPerturbations} in an image which we hope to exploit with the fluctuations of the frequencies in the spectrograms creating sharp textures, key peak frequencies, and smooth textures in others. The WGAN${_2}$ loss function in Equation \ref{lossFunctionWGAN_2} also leverages an Adam optimizer, independent in training but dependent on the output from $G$ in the WGAN${_1}$.

\begin{equation}
\label{lossFunctionWGAN_2}
L=\mathbb{E}_{\boldsymbol{y}\sim\mathbb{P}_{r}}[D_y(\boldsymbol{y})]-\mathbb{E}_{\boldsymbol{z\sim\mathcal{N}(0,1)}}[D(F(G(z)))] +\lambda\mathbb{E}_{\boldsymbol{\hat{y}}\sim\mathbb{P}_{\hat{y}}}[(||\nabla_{\boldsymbol{\hat{y}}}D(\boldsymbol{\hat{y}})||_{2}-1)^{2}]
\end{equation}

Again, notice the dependency of Equation \ref{lossFunctionWGAN_2} with the output of $G$ from WGAN${_1}$. We believe this helps contribute to TSGAN's ability to perform few shot learning. This is when little training data is needed in order to model realistic synthetic time series data. This is a trait of the architecture and helps in understanding that when these GANs are used together the downfall of "not enough data" is confronted. We see that in WGAN${_1}$ we are able to stabilize the training of the random input to create a 2D image which is essentially a heat map, meaning no real "natural/life-like" features are needed to generate good spectrogram images (unlike that of humans or animals) so good synthetic spectrogram images are created rather quickly. This ability for WGAN${_1}$ to generate good spectrogram images is the strength of the WGAN${_2}$ which is conditioned on those images. This is one of the first architectures of its kind to go from a image representation down to a time series representation purely in the architecture. Work performed by \cite{Gulrajani2017ImprovedGANs, Mescheder2018OnTraining} demonstrates that TSGAN attempts to stabilize its training with adding noise to the latent representation of the input to both generators and discriminators. We found this added robustness to our generated samples and challenged the discriminator so it could better train the generator and not over
\section{Experiments and Evaluation}
\label{S:3}

We perform our experiments on over 70 1D univariate time series data from the University of California Riverside (UCR) which has multiple subsets spanning various types of time series collections \cite{Bagnall2017TheAdvances}. This data set contains time series from sensor, spectrogram and image translated, device, motion, medical, simulated, traffic, speech, acoustic and audio data which are univariate or multivariate signals. We focus on the univariate data from the UCR data set and try to show the generation can be performed across these various types of signal types and refer the reader to the UCR data set archive for more information on each individual data set used. 

We choose data sets that contain 2-5 classes and we break those down into small, medium, and large training data sets. What this means is our TSGAN training data is the concatenation of the original UCR training and testing data. Our split is 0-499 signals as small training data sets, 500-1000 signals as medium training data sets, and anything greater than 1000 as large training data sets. We use this split to show TSGAN's ability to perform few shot learning for generation and hope to standardize these data sets for few shot 1D generation experiments. 

\subsection{Metrics}
Metrics to show quantitatively how well GANs are producing real like synthetic data is still an ongoing research area \cite{Barratt2018AScore, Borji2019ProsMeasures, Im2018QuantitativelyTraining, Lucic2018AreStudy}. The issue with many of these metrics in relation to our work is the majority of the GAN work has been in the 2D domain which means many of the metrics are biased for images. For example, looking at a popular metric for GAN generation in 2D is the Inception Score which is a measure of how realistic a GAN's output is correlating to human evaluation \cite{Salimans2016ImprovedGans}. This metric scores if the generated images have variety and each image looks like what it's suppose to look like. The issue with this metric for our TSGAN is Inception Score is only for images (2D) and is found by using a pretrained CNN (InceptionV3) trained on ImageNet \cite{Deng2009ImageNet:Database, Szegedy2017Inception-v4Learning}.

This means our 1D generations can't use the same metric for realism that is used in 2D and hugely in part because there is no base line standard data set in 1D like ImageNet in 2D. Another large issue in 1D compared to 2D is the human perception piece in how well the GAN is doing. For instance, many or most people have seen a dog, cat, or car and know what those objects are suppose to look like. This is not the same when looking at ECG signals or motion events to the point where subject mater expertise is needed to know if the signal is coming back correctly. One more issue with 1D data generation is the real life application of the signal; for example, if music is generated one can save off the generated sample as a playable file and then humans can listen to see if the generation sounds like music. This is not the same for signals like ECG or star light curves, all of which would require lengthy data collection and subject matter expertise. These issues and lack of community research make performance evaluation hard for our methodology in the 1D domain. 

Due to these, we look to first express our performance with an adjusted 1D Fréchet Inception Score (FID). The FID is supposed to improve on the Inception Score by actually comparing the statistics of generated samples to real samples, instead of evaluating generated samples in a vacuum \cite{Heusel2017GansEquilibrium}. The difference being in 1D we do not have a standardized data set similar to ImageNet, so instead we calculate a new 1D FID score independent for each of the data sets tested from the UCR repository. The lower the FID score corresponds to more similar real and generated samples as measured by the distance between their activation distributions.To do this score, we use a simple fully  convolutional network (FCN) trained on each data set independently \cite{Wang2017TimeBaseline}. The lower the 1D FID score the better. We find in the 1D domain and with a different 1D FID for each data set, there is no good characterization on the "perfect" FID score. Ideally, if it was a standardized data set incorporating all 1D time series signals, similar to a data set like ImageNet, then the ideal score would be zero. What we find in our experiments is even when signal are generated well visually, they might still not have FID scores approaching zero. However, the better visual results do correlate to the lesser of the FID scores. 

Looking at many of the works in Section \ref{relatedWorks}, there is a strong presence of using classification accuracy for a metric; i.e using the new synthetic data in some sort of classification task. With this in mind, we look at a classification accuracy metric based on two different scenarios:
train on synthetic data, test on real and train on real data, test on synthetic (TSTR and TRTS) \cite{Esteban2017Real-valuedGans}. We train a FCN algorithm on the generated samples and
test the accuracy on the real data set and then flip the training/testing, training the algorithm on the real data and testing it on the generated samples. We hope to see correlating accuracy compared to the performance of the FCN alone on the UCR data set. 

\section{Results}
\label{S:4}
We split the results discussion into two parts: first, qualitative visual results where we look at both the FID scores and some actual generated time series data. Second, we will look at the TSTR/TRTS scores where we hope to find a quantitative representation on how well our proposed method generates time series signals for real machine learning applications. 

\begin{figure}[h!]
    \begin{subfigure}{0.32\textwidth}
        \includegraphics[scale=.6]{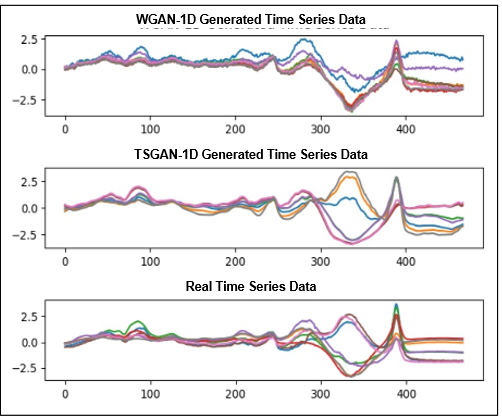}
        \caption{Class 1}
        \label{fig:subim1}
    \end{subfigure}
    \begin{subfigure}{0.32\textwidth}
        \includegraphics[scale=.6]{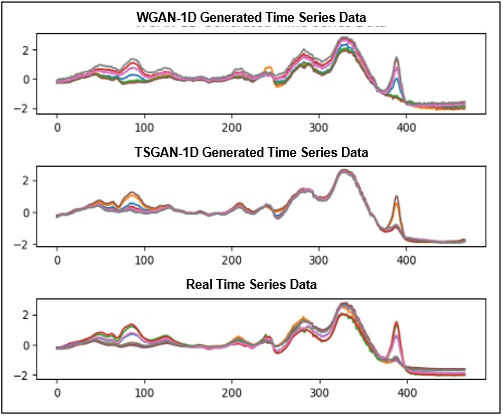}
        \caption{Class 2}
        \label{fig:subim2}
    \end{subfigure}
    \begin{subfigure}{0.32\textwidth}
        \includegraphics[scale=.6]{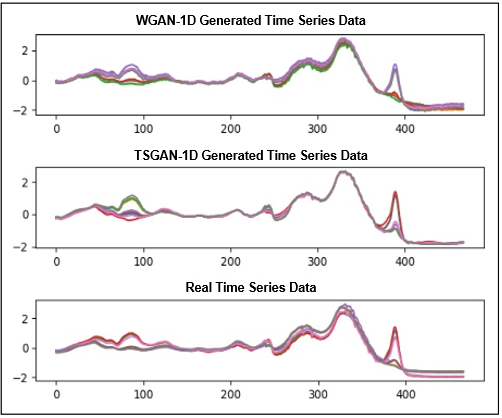}
        \caption{Class 3}
        \label{fig:subim3}
    \end{subfigure}
    \begin{subfigure}{0.32\textwidth}
        \includegraphics[scale=.6]{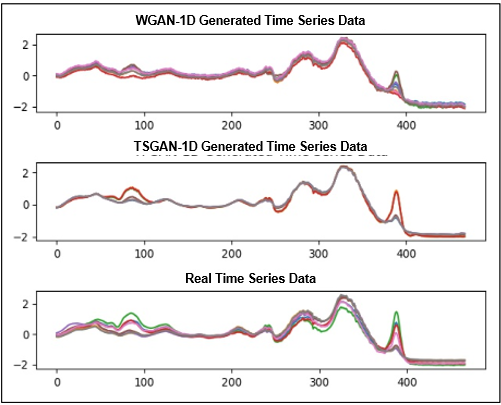}
        \caption{Class 4}
        \label{fig:subim2}
    \end{subfigure}
    \begin{subfigure}{0.32\textwidth}
        \includegraphics[scale=.6]{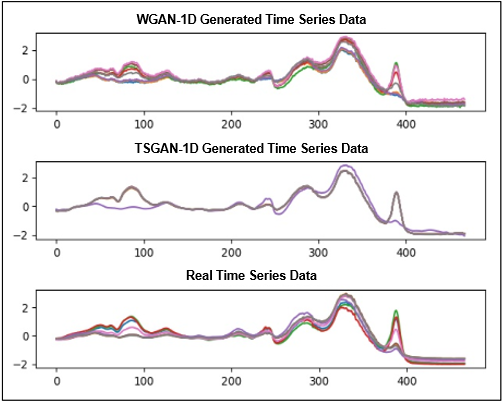}
        \caption{Class 5}
        \label{fig:subim2}
    \end{subfigure}
\caption{Results from testing on "Beef" data sets.  Each of the 5 classes is shown in subfigures (a)-(e).  The WGAN results are illustrated in the top sub-image, with the second sub-image showing TSGAN results, and the bottom sub-image is the REAL data.}
\label{fig:Beef}
\end{figure}

\begin{figure}[h!]
    \begin{subfigure}{0.49\textwidth}
        \includegraphics[scale=.5]{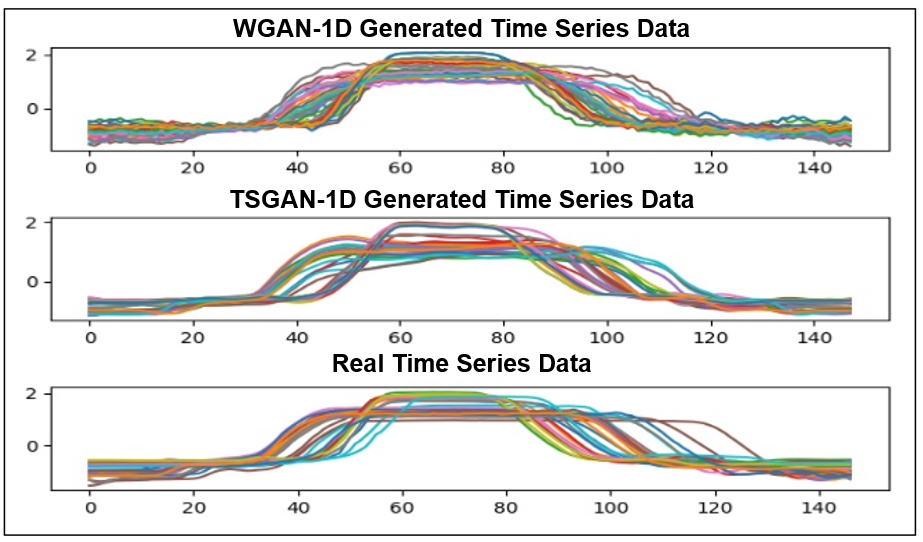}
        \caption{Class 1}
        \label{fig:subim1}
    \end{subfigure}
    \begin{subfigure}{0.49\textwidth}
        \includegraphics[scale=.5]{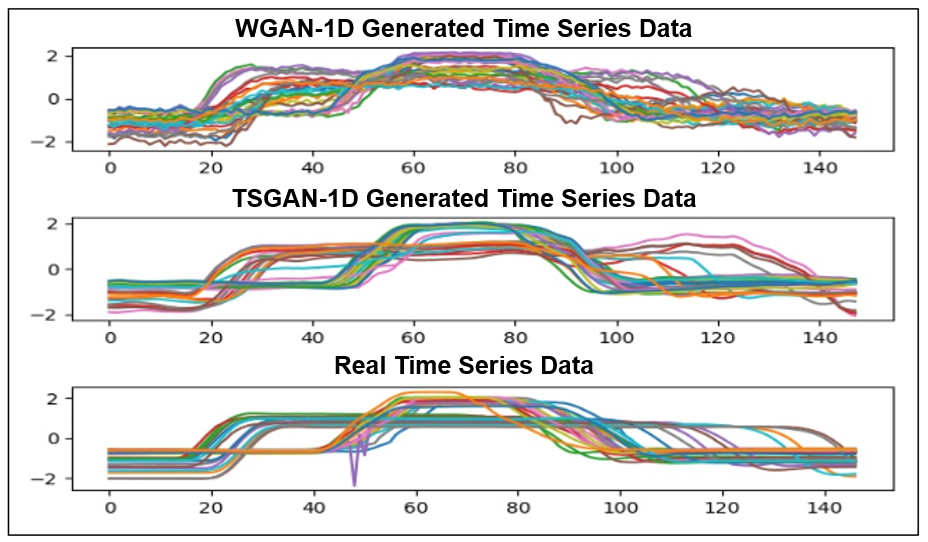}
        \caption{Class 2}
        \label{fig:subim2}
    \end{subfigure}
    \caption{Results from testing on "Gun Point" data sets.  Each of class is shown in subfigures (a) and (b).  The WGAN results are illustrated in the top sub-image, with the second sub-image showing TSGAN results, and the bottom sub-image is the REAL data.}
\label{fig:Gunpoint}
\end{figure}

\begin{figure}[h!]
    \begin{subfigure}{0.32\textwidth}
        \includegraphics[scale=.6]{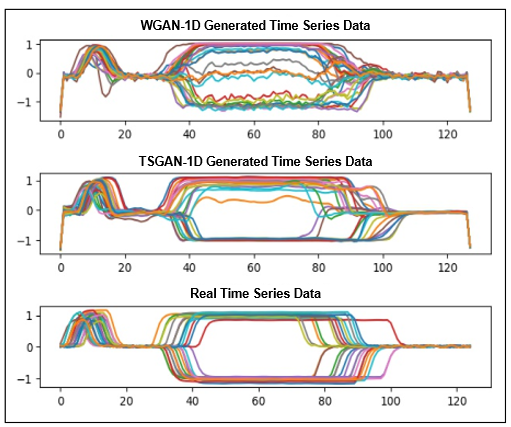}
        \caption{Class 1}
        \label{fig:subim1}
    \end{subfigure}
    \begin{subfigure}{0.32\textwidth}
        \includegraphics[scale=.6]{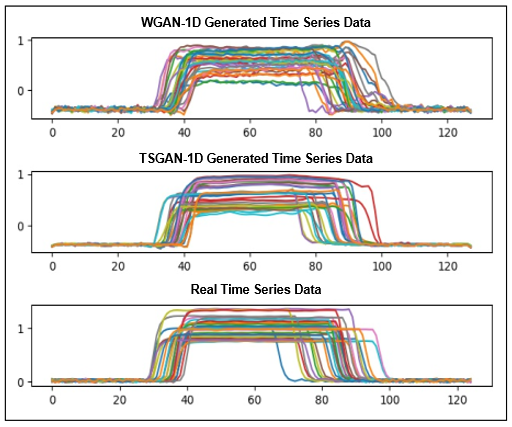}
        \caption{Class 2}
        \label{fig:subim2}
    \end{subfigure}
    \begin{subfigure}{0.32\textwidth}
        \includegraphics[scale=.6]{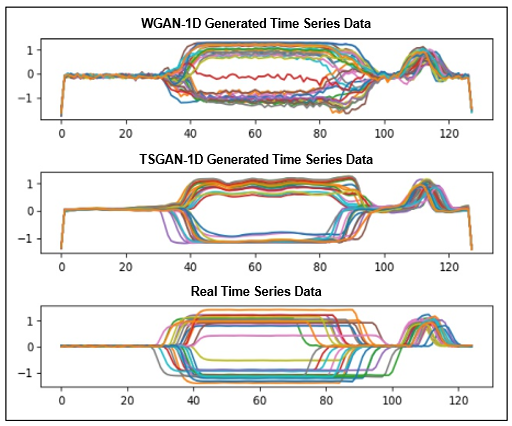}
        \caption{Class 2}
        \label{fig:subim2}
    \end{subfigure}
\caption{Visual comparison of TSGAN results on BME data sets. Each class is shown of the data sets, WGAN results are on top followed by TSGAN and then the real data on the bottom.}
\label{fig:BME}
\end{figure}

\begin{figure}[h!]
    \begin{subfigure}{0.32\textwidth}
        \includegraphics[scale=.6]{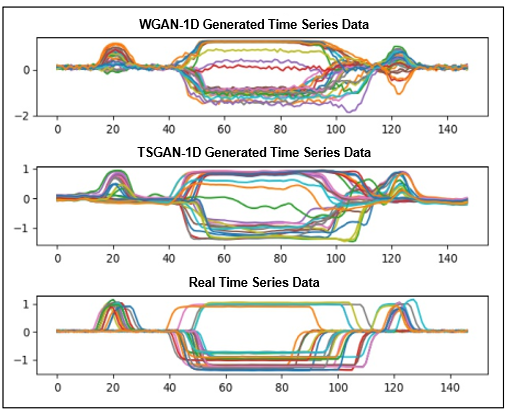}
        \caption{Class 1}
        \label{fig:subim1}
    \end{subfigure}
    \begin{subfigure}{0.32\textwidth}
        \includegraphics[scale=.6]{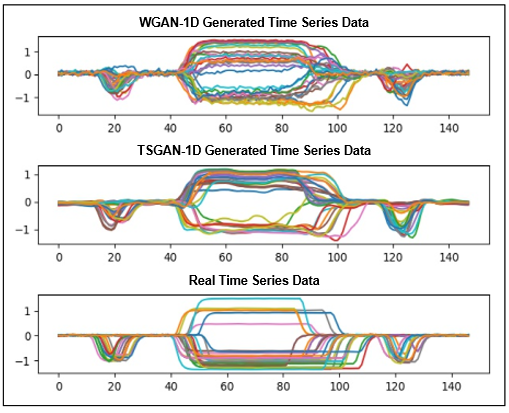}
        \caption{Class 2}
        \label{fig:subim2}
    \end{subfigure}
    \begin{subfigure}{0.32\textwidth}
        \includegraphics[scale=.6]{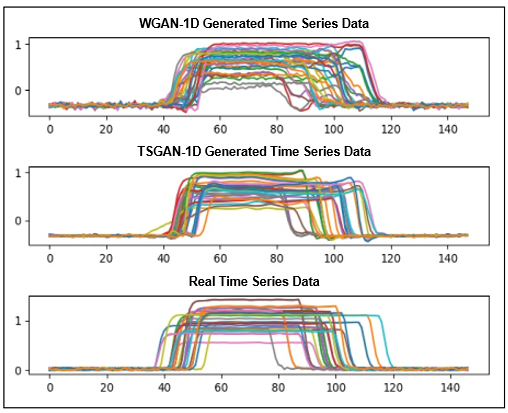}
        \caption{Class 2}
        \label{fig:subim2}
    \end{subfigure}
\caption{Visual comparison of TSGAN results on UMD data sets. Each class is shown of the data sets, WGAN results are on top followed by TSGAN and then the real data on the bottom.}
\label{fig:UMD}
\end{figure}

\begin{figure}[h!]
    \begin{subfigure}{0.49\textwidth}
        \includegraphics[scale=.5]{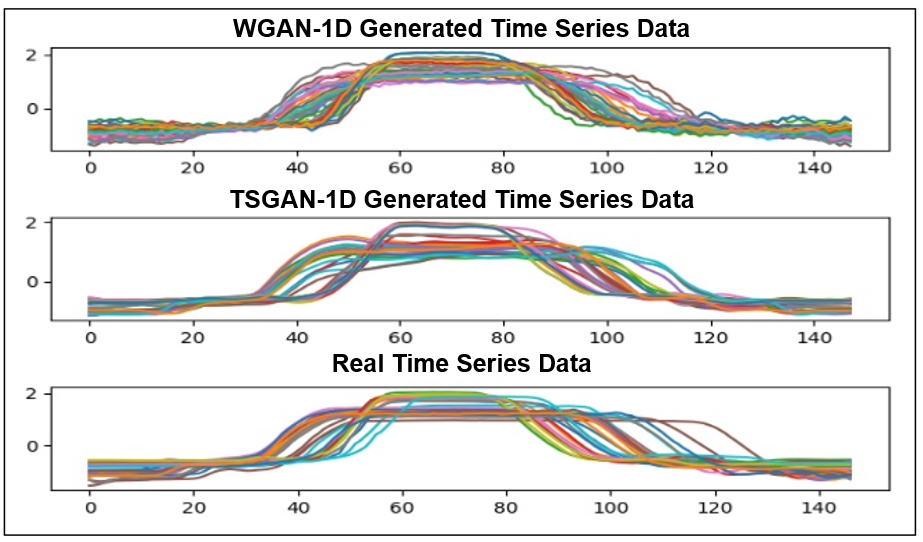}
        \caption{Class 1}
        \label{fig:subim1}
    \end{subfigure}
    \begin{subfigure}{0.49\textwidth}
        \includegraphics[scale=.5]{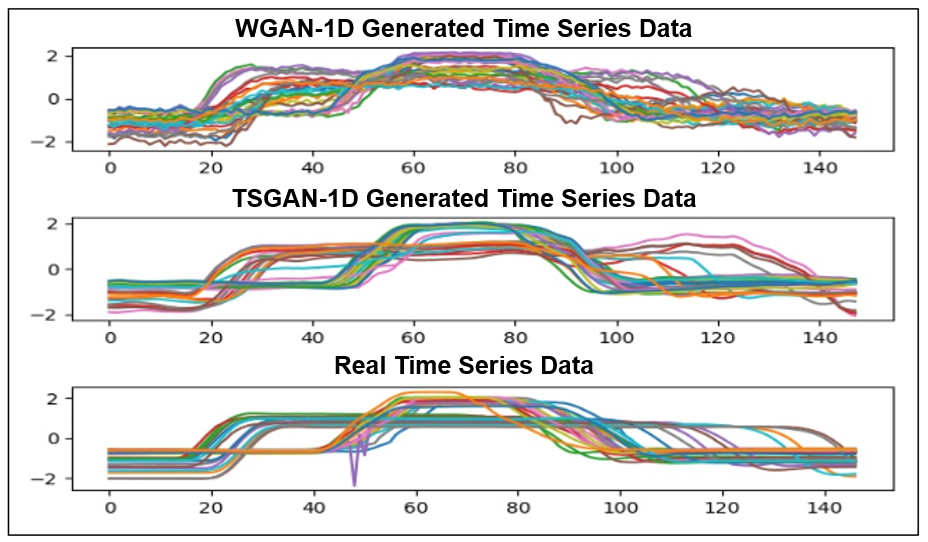}
        \caption{Class 2}
        \label{fig:subim2}
    \end{subfigure}
\caption{Results from testing on "Gun Point" data sets.  Each of class is shown in subfigures (a) and (b).  The WGAN results are illustrated in the top sub-image, with the second sub-image showing TSGAN results, and the bottom sub-image is the REAL data.}
\label{fig:Gunpoint}
\end{figure}

\begin{figure}[h!]
    \begin{subfigure}{0.49\textwidth}
        \includegraphics[scale=.6]{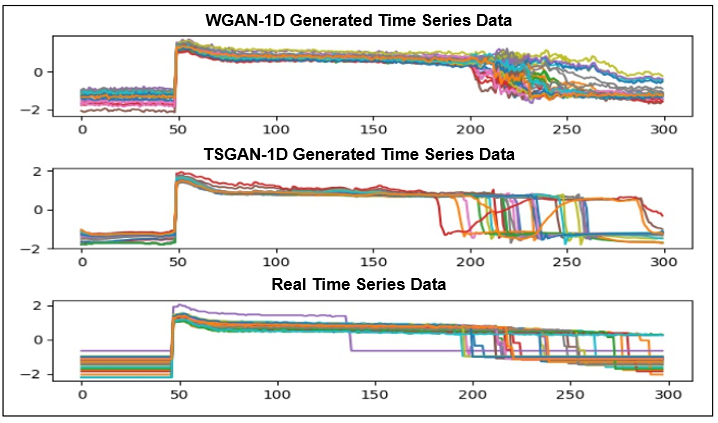}
        \caption{Class 1}
        \label{fig:subim1}
    \end{subfigure}
    \begin{subfigure}{0.49\textwidth}
        \includegraphics[scale=.6]{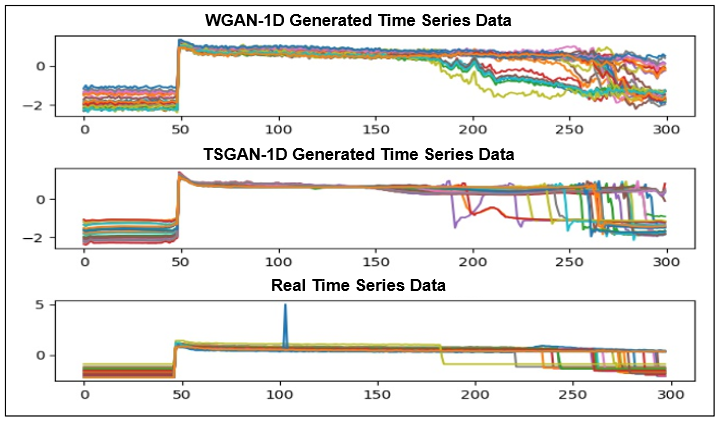}
        \caption{Class 2}
        \label{fig:subim2}
    \end{subfigure}
\caption{Visual comparison of TSGAN results on Freezer Regular data sets. Each class is shown of the data sets, WGAN results are on top followed by TSGAN and then the real data on the bottom. }
\label{fig:FreezerReg}
\end{figure}

\begin{figure}[h!]
    \begin{subfigure}{0.49\textwidth}
        \includegraphics[scale=.6]{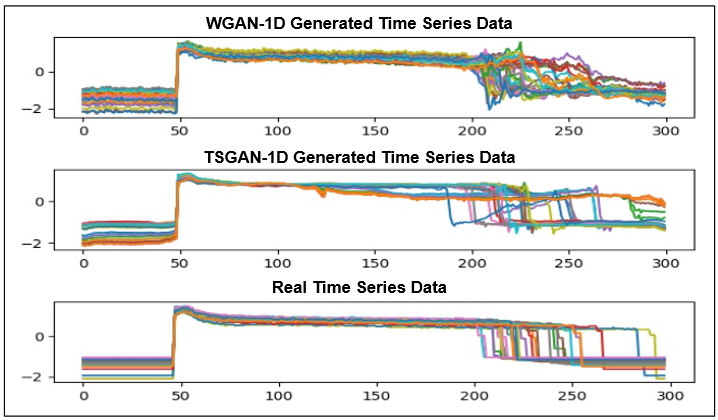}
        \caption{Class 1}
        \label{fig:subim1}
    \end{subfigure}
    \begin{subfigure}{0.49\textwidth}
        \includegraphics[scale=.6]{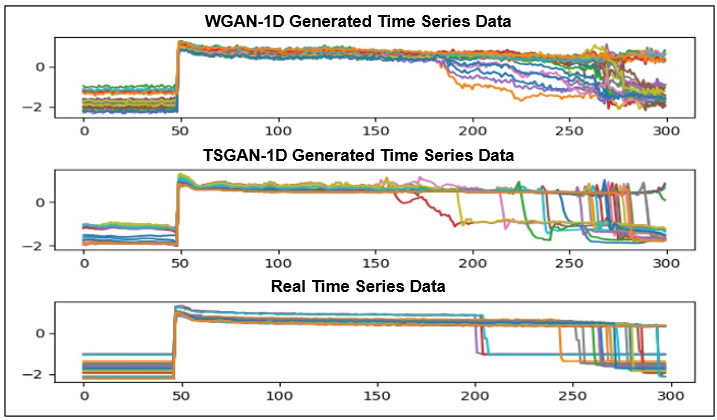}
        \caption{Class 2}
        \label{fig:subim2}
    \end{subfigure}
\caption{Visual comparison of TSGAN results on Freezer Small data sets. Each class is shown of the data sets, WGAN results are on top followed by TSGAN and then the real data on the bottom. }
\label{fig:FreezerSmall}
\end{figure}

\begin{figure}[h!]
    \begin{subfigure}{0.49\textwidth}
        \includegraphics[scale=.6]{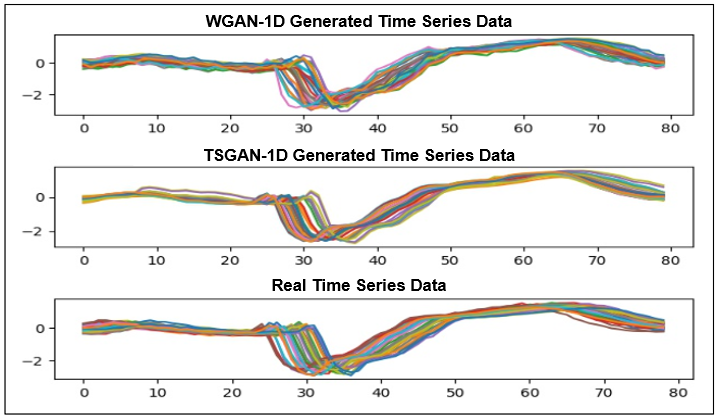}
        \caption{Class 1}
        \label{fig:subim1}
    \end{subfigure}
    \begin{subfigure}{0.49\textwidth}
        \includegraphics[scale=.6]{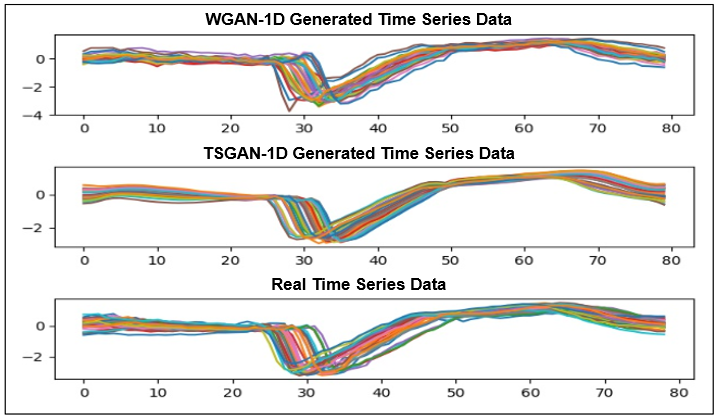}
        \caption{Class 2}
        \label{fig:subim2}
    \end{subfigure}
\caption{Visual comparison of TSGAN results on Two Lead ECG data sets.  Each class is shown of the data sets, WGAN results are on top followed by TSGAN and then the real data on the bottom.}
\label{fig:ecg}
\end{figure}

Experimental results using TSGAN to generate time series data will show its ability in generating refined and realistic time series signals. Visually, a few data sets can be seen in Figure \ref{fig:Beef} - \ref{fig:ecg}.  We use the FID score in order to gauge quantitatively how well our visual comparison is. For the first data set shown, Beef in Figure \ref{fig:Beef}, our method achieves a FID at 207.1 while WGAN receives a 638.8, showing more than three times better quantitative results then WGAN. We see on this data set that WGAN and TSGAN visually perform well though, showing why a metric is so valuable in comparing these synthetic time series data. The Gun Point data set is an interesting data set for comparison. Looking at it visually we see that TSGAN signals look much better than that of WGAN's generation. When looking at comparison of the signals using our other metric (TSTR/TRTS) would seem both WGAN and TSGAN are equally able. However, when looking at our FID score we can see our method achieves a 4.778 FID score while WGAN's performance is FID score of 39.95. This again is why multiple metric are needed when comparing these generations.   

Looking at Figure \ref{fig:BME} and  Figure \ref{fig:UMD}, the TSGAN obtains a 20.61 on and 41.05 respectively, which again outperforms WGAN which achieves a 53.27 and 1694 respectively. What we see with these two data sets is TSGAN ability to perform generation on sharp rigid changes in the signal an do so smoothly. WGAN on the other hand is extremely noisy and unable to perform the sharp transitions. 

These four data sets shown above are data sets that contain a small amount of training samples for our TSGAN training. This solidifies TSGAN's ability to perform few shot learning. We believe the architecture of TSGAN is what gives its ability of few shot learning. With the generation of spectrograms images with the first WGAN (exploiting WGAN's ability to 2D images well) we condition the second WGAN to perform even better generation. This differs from the original WGAN taking a random input vector to model time series, where that random vector has no influential information to help generate time series. 

Looking at the other four data sets we can see again TSGAN outperforms WGAN in the FID scores. These four data sets have large data sets for training the GANS. Similar to the visual comparisons above, we see TSGAN generates well when there are sharp transitions and peaks in time steps. WGAN fails miserable and looks noisy and the overall structure on these data sets do not resemble that of the original data sets. 

Overall in the 70 data sets we experimented on, TSGAN performed better in the FID visualization metric better on 56 data sets. Of those, 15 of the 16 large training sets, 14 of the 18 medium training sets, and 27 of the 36 small training data sets. Of these nine small data sets WGAN outperforms TSGAN the data sets are extremely difficult for differentiation for the classifier, which leads to the assumption that though WGAN outperforms TSGAN on these data sets the actual generation was poor for both methods. We can infer through TSGAN's performance in our experiments that the conditioned generator for the second WGAN allows for better visual results in the time series data (both backed by the FID score and human perception). This can possibly be explained by the condition being instilled on the input of the second WGAN is actually learned from the first WGAN, instead of just fed to the WGAN. This learning of the conditioned based off the original time series data must strengthen the input space and allow for better synthetic generation, accounting for both temporal and frequency characteristics in the time series signature.

For classification accuracy (TSTR and TRTS) we compare our accuracy
with the accuracy of training on real and testing on real (TRTR)  results demonstrated in Esteban et al's paper for medical time series generation \cite{Esteban2017Real-valuedGans}. We re-train the same 
FCN architecture on each subset of the UCR data set shown above and run each training of the FCN 1000 epochs \cite{Wang2017TimeBaseline}.

We keep the training and testing size of each class the same as is seen in the UCR data set. To summarize, if the training data on a class is 50 signals and the testing on a class is 100 signals in the real UCR data subset, we use 50 real signals and 100 synthetic signals to do TRTS and 50 synthetic signals and 100 real signals to do TSTR. To note, the accuracy could be increased if there was an unlimited amount of synthetic data allowed to use for TSTR training phase, since our TSGAN could model more realistic class
signals for the training phase. TRTS metric shows the TSGAN has the ability to generate realistic data and demonstrates the ability to generate data that could be used for real applications while the TSTR metric fortifies that TSGAN is creating samples that look realistic and has features similar to the original data set.

\begin{figure}[H]
    \centering
    \includegraphics[scale=.8]{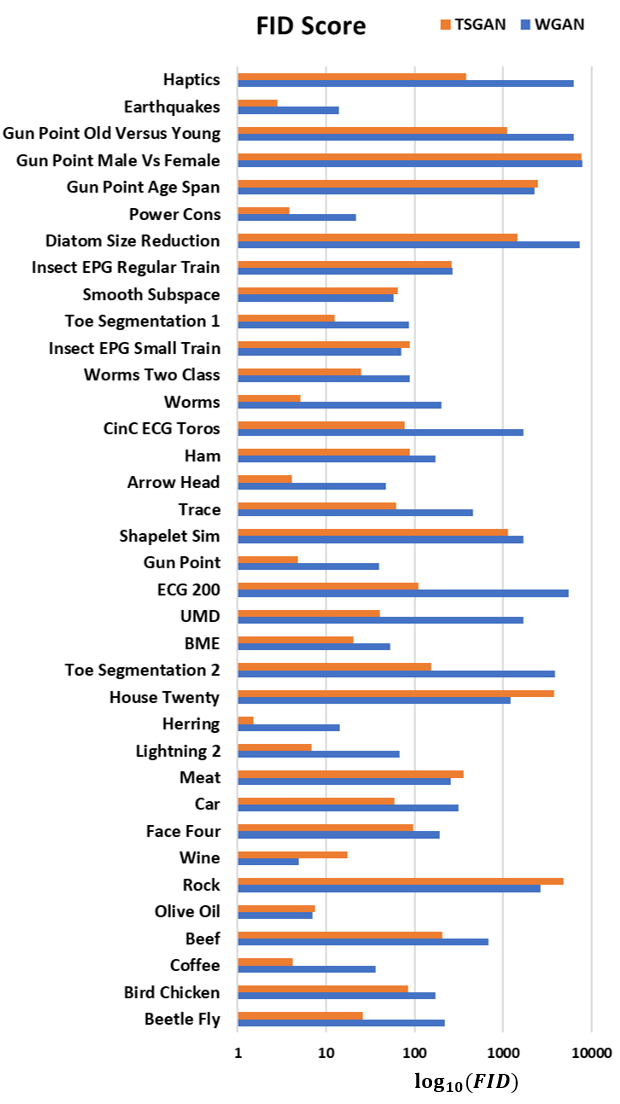}
    \caption{WGAN and TSGAN FID score comparison on small sized data sets.} 
    \label{smallfigfid}
\end{figure}

\begin{figure}[H]
    \centering
    \includegraphics[scale=.8]{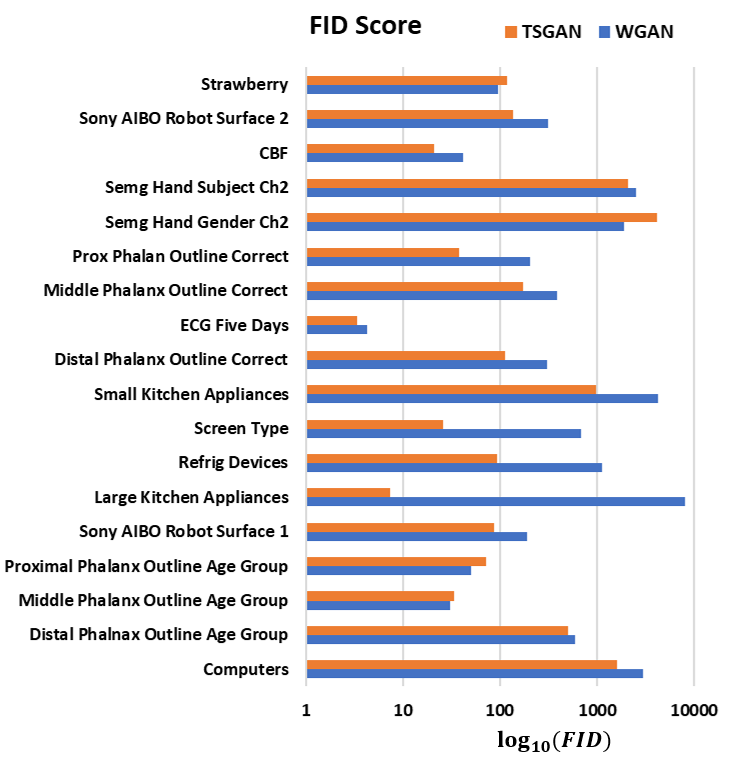}
    \caption{WGAN and TSGAN FID score comparison on medium sized data sets.} 
    \label{medfigfid}
\end{figure}

\begin{figure}[H]
    \centering
    \includegraphics[scale=.7]{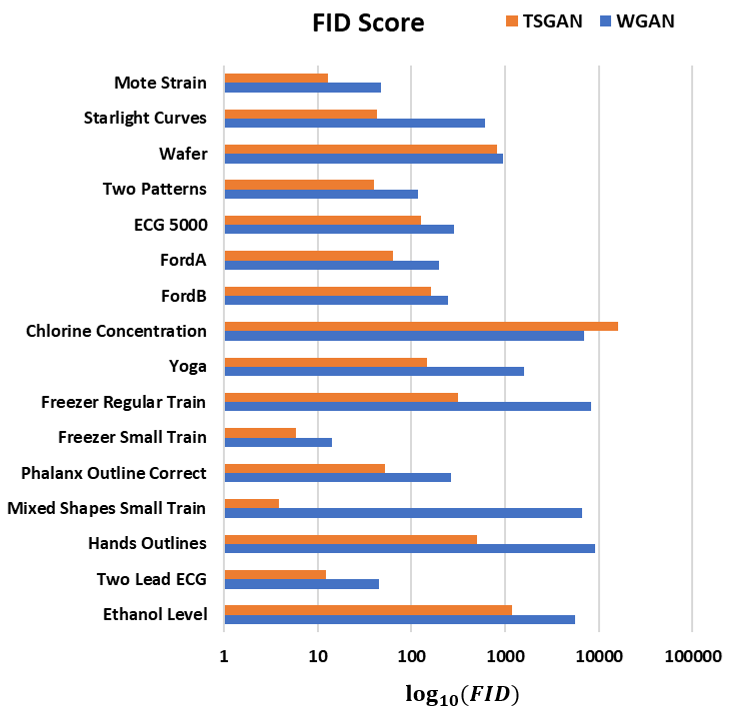}
    \caption{WGAN and TSGAN FID score comparison on large size data sets.} 
    \label{largefid}
\end{figure}

We consider both metrics because TRTS accuracy will not be hindered in the case that our GAN suffers to mode collapse but TSTR will diminish significantly. The results of this test can be seen in Table \ref{tab:UCR_dataset_small} - Table \ref{table:UCR_dataset_large}, and FID scores in Figure \ref{smallfigfid} - Figure \ref{largefid}, showing the different small, medium, and large generation training sets.
Our results for TSGAN in accuracy of TRTS and TSTR correlate well with the real accuracy of the  classification algorithm
on the standard data set (TRTR) and outperforms the WGAN. We do have several data sets where we suffer from mode collapse and for these data sets we believe TSGAN and WGAN both suffered which leads to the belief that the internal WGAN hyperparameters were in need of fine tuning. This also gives good insight that a hyperparameter search might help in the generation of time series data, something to look into for future research.  These results show a measure of similarity our generated time series data has compared to the original time series data. 

What this explains is the synthetic data being generated from TSGAN contains the same (or closely related) features which makes up the original time series data. This shows the new generated sequences from TSGAN carries the same temporal dynamics as the original data between variables across time. 
Comparing these metrics in accordance to the competition against WGAN isn't enough to get a good idea what's going on. To go one step further we look at comparing the metrics against what the FCN does on the actual training and testing on the data sets. This we ran just like the above mentioned experiments and record the accuracy in TRTR. What we hope to see is similar results across all three metrics which would mean the generated samples share the same learned features as in the regular data set's training/testing signals. 

In summary out of the 70 data sets used TSGAN outperforms WGAN on TRTS in 36 data sets, and ties the WGAN in 16 of the data sets. Interesting to look at is the ties, they are either 100\% accuracy or 0\% accuracy, the ladder being the same data sets we believe were mode collapse and the 100\% data sets where traditional machine learning algorithms also performed well. In TSTR, TSGAN outperforms the WGAN in 54 of the data sets while tying in 5 of the data sets. On those data sets in which WGAN performs better is within 1\% of TSGA's results, meaning they are closely related and within some correlation between each other. 

We see in our experimental results that TSGAN does outperform WGAN in the generation metrics we are using. We believe this is from TSGAN's ability to condition the input to the time series generation WGAN utilizing more information in the latent space than just a random vector. Not only does TSGAN generate internal features similar to real data, it also generates less noisy samples that have similar frequency oscillations compared to that of the WGAN, looking visually at the plots above. 

\begin{table}[h!]
\begin{footnotesize}
\centering
\begin{tabular}{|c|c|c||c|c||c|}
\hline
\textbf{\begin{tabular}[c]{@{}c@{}}Data Set\end{tabular}} &
\textbf{\begin{tabular}[c]{@{}c@{}}WGAN\\  TRTS\end{tabular}} &
\textbf{\begin{tabular}[c]{@{}c@{}}TFGAN\\  TRTS\end{tabular}} &
\textbf{\begin{tabular}[c]{@{}c@{}}WGAN \\ TSTR\end{tabular}} &
\textbf{\begin{tabular}[c]{@{}c@{}}TFGAN \\ TSTR\end{tabular}} &
\textbf{\begin{tabular}[c]{@{}c@{}}TRTR\end{tabular}} \\
\hline

\rowcolor[HTML]{ECF4FF} 
Beetle Fly &
  \textbf{100} &
  \textbf{100} &
  55 &
  \textbf{90} &
  \textit{85} \\ \hline
\rowcolor[HTML]{ECF4FF} 
\begin{tabular}[c]{@{}c@{}}Bird Chicken\end{tabular} &
  65 &
  \textbf{100} &
  90 &
  \textbf{75} &
  \textit{90} \\ \hline
\rowcolor[HTML]{ECF4FF} 
Coffee &
  \textbf{100} &
  \textbf{100} &
  \textbf{100} &
  \textbf{100} &
  \textit{100} \\ \hline
\rowcolor[HTML]{ECF4FF} 
Beef &
  6.67 &
  \textbf{96.67} &
  20 &
  \textbf{60} &
  \textit{53.33} \\ \hline
\rowcolor[HTML]{ECF4FF} 
Olive Oil &
  0 &
  0 &
  \textbf{40} &
  \textbf{40} &
  \textit{40} \\ \hline
\rowcolor[HTML]{ECF4FF} 
Rock &
  0 &
  0 &
  20 &
  \textbf{36} &
  \textit{18} \\ \hline
\rowcolor[HTML]{ECF4FF} 
Wine &
  \textbf{13} &
  0 &
  55.56 &
  \textbf{61.11} &
  \textit{55.56} \\ \hline
\rowcolor[HTML]{ECF4FF} 
Face Four &
  28.4 &
  \textbf{64.77} &
  65.22 &
  \textbf{84.22} &
  \textit{93.18} \\ \hline
\rowcolor[HTML]{ECF4FF} 
Car &
  \textbf{70} &
  50 &
  65 &
  \textbf{70} &
  \textit{86.67} \\ \hline
\rowcolor[HTML]{ECF4FF} 
Meat &
  \textbf{10} &
  0 &
  \textbf{88.33} &
  46.67 &
  \textit{68.33} \\ \hline
\rowcolor[HTML]{ECF4FF} 
Lighting2 &
  62.3 &
  \textbf{72.95} &
  70.49 &
  \textbf{72.21} &
  \textit{70.49} \\ \hline
\rowcolor[HTML]{ECF4FF} 
Herring &
  73.44 &
  \textbf{100} &
  46.88 &
  \textbf{65.62} &
  \textit{71.87} \\ \hline
\rowcolor[HTML]{ECF4FF} 
\begin{tabular}[c]{@{}c@{}}House Twenty\end{tabular} &
  \textbf{45.37} &
  36.13 &
  \textbf{57.98} &
  42.07 &
  \textit{57.98} \\ \hline
\rowcolor[HTML]{ECF4FF} 
\begin{tabular}[c]{@{}c@{}}Toe Segmentation 2\end{tabular} &
  28.46 &
  \textbf{83.08} &
  80.77 &
  \textbf{91.53} &
  \textit{90} \\ \hline
\rowcolor[HTML]{ECF4FF} 
BME &
  14 &
  \textbf{34} &
  80 &
  \textbf{82.67} &
  \textit{76} \\ \hline
\rowcolor[HTML]{ECF4FF} 
UMD &
  43.75 &
  \textbf{87.5} &
  97.22 &
  \textbf{97.91} &
  \textit{97.91} \\ \hline
\rowcolor[HTML]{ECF4FF} 
ECG200 &
  99 &
  \textbf{100} &
  82 &
  \textbf{82} &
  \textit{91} \\ \hline
\rowcolor[HTML]{ECF4FF} 
Gun Point &
  \textbf{100} &
  \textbf{100} &
  \textbf{100} &
  \textbf{100} &
  \textit{98.67} \\ \hline

\end{tabular}
\caption{TRTS, TSTR, TRTR accuracy on the UCR small data set with total samples ($<$500). 
\label{tab:UCR_dataset_small}}
\end{footnotesize}
\end{table}

% Medium size data set statistics
\begin{table}[H]
\begin{footnotesize}
\centering
\begin{tabular}{|c|c|c||c|c||c|}
\hline
\textbf{\begin{tabular}[c]{@{}c@{}}Data Set\end{tabular}} &
\textbf{\begin{tabular}[c]{@{}c@{}}WGAN\\  TRTS\end{tabular}} &
\textbf{\begin{tabular}[c]{@{}c@{}}TFGAN\\  TRTS\end{tabular}} &
\textbf{\begin{tabular}[c]{@{}c@{}}WGAN \\ TSTR\end{tabular}} &
\textbf{\begin{tabular}[c]{@{}c@{}}TFGAN \\ TSTR\end{tabular}} &
\textbf{\begin{tabular}[c]{@{}c@{}}TRTR\end{tabular}} \\
\hline

\rowcolor[HTML]{FFFFC7} 
Computers & 53.2 & \textbf{100} & 50.8 & \textbf{65.2} & \textit{78.8} \\ \hline

\rowcolor[HTML]{FFFFC7} 
\begin{tabular}[c]{@{}c@{}}Distal Phalnax Outline \\Age Group\end{tabular} &
  88.49 & \textbf{93.52} & \textbf{73.38} & 70.5 & \textit{73.38}\\ \hline

\rowcolor[HTML]{FFFFC7} 
\begin{tabular}[c]{@{}c@{}}Middle Phalanx Outline \\Age Group\end{tabular} &
  1.3 & \textbf{5.84} & 40.9 & \textbf{50} & \textit{60.39} \\ \hline

\rowcolor[HTML]{FFFFC7} 
\begin{tabular}[c]{@{}c@{}}Proximal Phalanx Outline \\Age Group\end{tabular} &
  \textbf{100} & \textbf{100} & \textbf{87.31} & 85.85 & \textit{77.07} \\ \hline

\rowcolor[HTML]{FFFFC7} 
\begin{tabular}[c]{@{}c@{}}Sony AIBO Robot \\Surface 1\end{tabular} &
  \textbf{42.76} & 40.76 & 89.51 & \textbf{92.84} & \textit{86.52} \\ \hline

\rowcolor[HTML]{FFFFC7} 
\begin{tabular}[c]{@{}c@{}}Large Kitchen Appl\end{tabular} &
  \textbf{99.47} & 89.87 & 59.47 & \textbf{74.93} & \textit{87.73} \\ \hline

\rowcolor[HTML]{FFFFC7} 
\begin{tabular}[c]{@{}c@{}}Refrig Devices\end{tabular} &
  89.33 & \textbf{96} & 33.06 & \textbf{42.4} & \textit{46.4} \\ \hline

\rowcolor[HTML]{FFFFC7} 
\begin{tabular}[c]{@{}c@{}}Screen Type\end{tabular} &
  29.33 & \textbf{41.07} & 52 & \textbf{56.8} & \textit{65.07} \\ \hline

\rowcolor[HTML]{FFFFC7} 
\begin{tabular}[c]{@{}c@{}}Small Kitchen Appl\end{tabular} &
  \textbf{1.6} & 0 & 56.53 & \textbf{64.01} & \textit{80.8} \\ \hline

\rowcolor[HTML]{FFFFC7} 
\begin{tabular}[c]{@{}c@{}}Distal Phalanx Outline \\Correct\end{tabular} &
  \textbf{80.8} & 59.78 & \textbf{68.84} & 64.86 & \textit{77.54} \\ \hline

\rowcolor[HTML]{FFFFC7} 
\begin{tabular}[c]{@{}c@{}}ECG Five Days\end{tabular} &
  70.03 & \textbf{75.85} & 92.56 & \textbf{98.84} & \textit{99.31} \\ \hline

\end{tabular}
\caption{TRTS, TSTR, TRTR accuracy scores on the UCR mediunm data sets with samples (500$-$1000). }
\label{table:UCR_dataset_medium}
\end{footnotesize}
\end{table}

% Large size data set statistics
\begin{table}[H]
\begin{footnotesize}
\centering
\begin{tabular}{|c|c|c||c|c||c|}
\hline
\textbf{\begin{tabular}[c]{@{}c@{}}Data Set\end{tabular}} &
\textbf{\begin{tabular}[c]{@{}c@{}}WGAN\\  TRTS\end{tabular}} &
\textbf{\begin{tabular}[c]{@{}c@{}}TFGAN\\  TRTS\end{tabular}} &
\textbf{\begin{tabular}[c]{@{}c@{}}WGAN \\ TSTR\end{tabular}} &
\textbf{\begin{tabular}[c]{@{}c@{}}TFGAN \\ TSTR\end{tabular}} &
\textbf{\begin{tabular}[c]{@{}c@{}}TRTR\end{tabular}} \\
\hline

\rowcolor[HTML]{FFCCC9} 
Ethanol Level & \textbf{100} & 0 &  24.8 & \textbf{30.2} & \textit{56.4} \\ \hline

\rowcolor[HTML]{FFCCC9} 
Two Lead ECG & 90.6 & \textbf{100} & 99.47 & \textbf{99.82} & \textit{99.91} \\ \hline

\rowcolor[HTML]{FFCCC9} 
Hand Outlines & 98.67 & \textbf{100} & 64.05 & \textbf{65.67} & \textit{75.13} \\ \hline

\rowcolor[HTML]{FFCCC9} 
Mixed Shapes Small Train & 0 & \textbf{85.65} & 55.54 & \textbf{82.64} & \textit{90.8} \\ \hline

\rowcolor[HTML]{FFCCC9} 
Phal Outl Correct & \textbf{64.45} & 50.35 & 73.89 & \textbf{76.92} & \textit{78.08} \\ \hline

\rowcolor[HTML]{FFCCC9} 
Freezer Small Train & 0 & \textbf{65.44} & 50.7 & \textbf{75.96} & \textit{71.4} \\ \hline

\rowcolor[HTML]{FFCCC9} 
Freezer Regular Train & \textbf{100} & \textbf{100} & 50 & \textbf{50.1} & \textit{99.4} \\ \hline

\end{tabular}
\caption{TRTS, TSTR, TRTR accuracy scores on the UCR large data sets with samples ($>$1000). }
\label{table:UCR_dataset_large}
\end{footnotesize}
\end{table}

\section{Conclusion}
The merging of time series data and deep learning is crucially important in the advancement of technology that impacts human life. With this in mind, it is also important to be able to leverage deep learning in multitudes of time series applications, even if data is restricted or limited. Typically, addressing this part of small data sets would be where generation software comes into play, or in some instances, a data collection where mass amounts of data are collected regarding the situation or application desired.  In situations where this is not an option, many machine learning algorithms run into problems with time series data, but this is where we look to advance the area with our TSGAN. In this paper we have shown TSGAN has the ability to generate realistic synthetic 1D signals across a multitude of data types: sensor, medical, simulated, motion, and other within human perceptual desire.

Our TSGAN also performs well in the task of being able to have its synthetic data used in real applications, such as creating a classifier based on the synthetic data and testing on the real data (and vice versa).  TSGAN's overall accuracy on all 70 data sets is roughly 11\% better than WGAN in TRTS, approximately 15\% better in TSTR, and only 4\% lower than the accuracy of TRTR, showing the data generated by TSGAN is closely related to the original data set.  Since there is little literature with 1D signals in time series domain being the main topic of generation, it is safe to say TSGAN has shown state of the art results in both time series generation and generation of signals with little to no training data (i.e. few shot).

\bibliographystyle{unsrt}  


\bibliography{references.bib}

\end{document}